\documentclass[nohyperref]{article}

\usepackage{microtype}
\usepackage{graphicx}
\usepackage{subfigure}
\usepackage{booktabs} 

\usepackage{hyperref}

\usepackage[accepted]{icml2022}

\usepackage{amsmath}
\usepackage{amssymb}
\usepackage{mathtools}
\usepackage{amsthm}

\usepackage[capitalize,noabbrev]{cleveref}

\usepackage{graphicx}
\usepackage{amsmath}
\usepackage{amssymb}
\usepackage{booktabs}

\usepackage{color}
\usepackage{multirow}
\usepackage{comment}
\usepackage{bm}

\usepackage{booktabs}

\usepackage{algorithm}
\usepackage{algorithmic}
\usepackage{xr}
\usepackage{amssymb}%
\usepackage{amsmath}
\usepackage{wrapfig}
\usepackage{graphicx}

\usepackage{bbm}
\usepackage{verbatim}

\usepackage{array}
\newcolumntype{H}{>{\setbox0=\hbox\bgroup}c<{\egroup}@{}}

\theoremstyle{plain}

\theoremstyle{definition}

\theoremstyle{remark}

\usepackage[textsize=tiny]{todonotes}

\icmltitlerunning{DAAS: Differentiable Architecture and Augmentation Policy Search}

\begin{document}

\twocolumn[
\icmltitle{DAAS: Differentiable Architecture and Augmentation Policy Search}



\icmlsetsymbol{equal}{*}

\begin{icmlauthorlist}
\icmlauthor{Xiaoxing Wang}{sjtu}
\icmlauthor{Xiangxiang Chu}{meituan}
\icmlauthor{Junchi Yan}{sjtu}
\icmlauthor{Xiaokang Yang}{sjtu}
\end{icmlauthorlist}

\icmlaffiliation{sjtu}{Department of CSE, and MoE Key Lab of Artificial Intelligence, AI Institute, Shanghai Jiao Tong University}
\icmlaffiliation{meituan}{Meituan}

\icmlcorrespondingauthor{Junchi Yan}{yanjunchi@sjtu.edu.cn}

\icmlkeywords{Machine Learning, ICML}

\vskip 0.3in
]



\printAffiliationsAndNotice{}  

\begin{abstract}
Neural architecture search has been an active direction of AutoML, aiming to explore efficient network structures. The discovered architectures are evaluated by training with fixed data augmentation policies. However, recent works on auto-augmentation show that the suited augmentation policies can vary over different structures. Therefore, this work considers the possible coupling between neural architectures and data augmentation and constructs a bi-level optimization for the joint search by refining the optimization target for NAS and AA. An effective differentiable search algorithm, named DAAS, is proposed based on Gumbel-softmax reparameterization (for NAS) and policy gradient (for AA). In particular, we point out the biased gradient approximation in the prior differentiable AA method and propose a novel and efficient method based on the policy gradient to overcome the above problem. 
Our approach can simultaneously search for efficient architecture and augmentation policies in 1 GPU-day and achieves 97.91\% accuracy on CIFAR-10 and 76.6\% Top-1 accuracy on the ImageNet dataset, showing the outstanding performance of our search method.
\end{abstract}

\section{Introduction}
AutoML aims to automatically construct and train machine learning models without human participation, of which Auto-augmentation (AA) and Neural Architecture Search (NAS) are two popular directions. A series of mechanisms have been designed, including reinforcement learning \cite{AA,zoph2017learning}, evolutionary algorithm \cite{real2019regularized}, Bayesian optimization \cite{fast-aa,bananas}, and gradient-based methods \cite{dada,liu2018darts}. 
These works are dedicated to either AA or NAS, but few explore joint searching for data augmentation policies and neural architectures.

\begin{figure}[tb]
	\centering
 \subfigure[Normal cell]{
		\includegraphics[width=0.35\columnwidth]{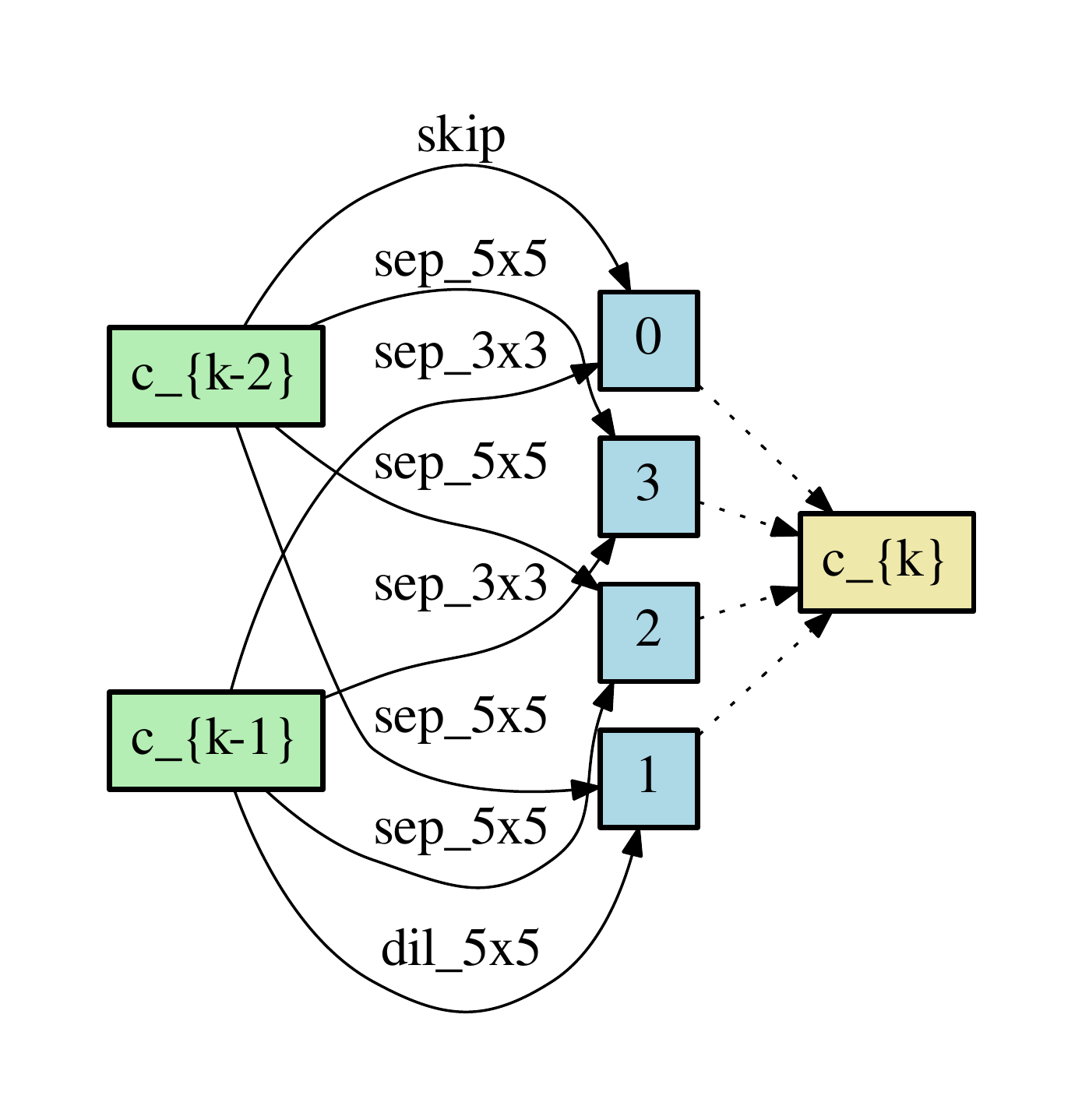}
    \label{fig:topology_normal}}
\vspace{-5pt}
\subfigure[Reduction cell]{
		\includegraphics[width=0.58\columnwidth]{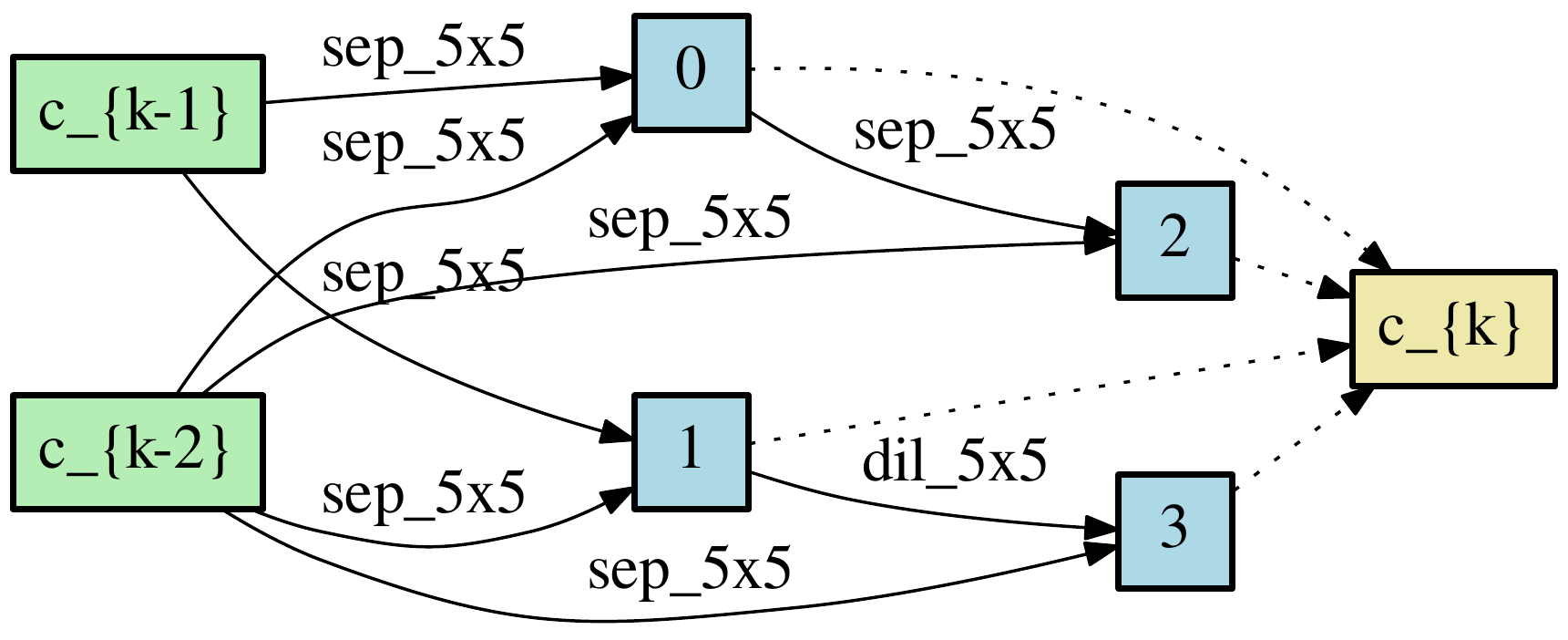}
    \label{fig:topology_reduce}}
    \vspace{-6pt}
	\caption{Motivating example: the architecture of (a) normal and (b) reduction cells discovered by our method. We observe that cells are dominated by 5x5 separable convolutions, which may not be well recommended in NAS literature. However, its accuracy can be improved by 0.5\% by suited data augmentation policy, achieving 97.91\% on CIFAR-10. Comparison of the default augmentation policies and our searched ones is reported in Table~\ref{tab:compare_only_nas}.}
	\label{fig:topology}
	\vskip -10pt
\end{figure}

However, there are connections between data augmentation policies and neural architectures. On the one hand, the performance of the architecture searched by NAS can be further improved by proper data augmentation. Fig.~\ref{fig:topology} shows an architecture searched by our method, which attains 97.42\% accuracy on CIFAR-10~\cite{krizhevsky2009learning} under the default data augmentation policy of DARTS~\cite{liu2018darts} but achieves 97.91\% under our searched policies;
On the other hand, the optimal augmentation policies for different architectures may also vary (see Sec.~\ref{subsec:ablation_study}).
We argue and show that (see experiments): 1) It is beneficial to jointly search for data augmentation policies and network architectures. 2) The vanilla evaluation metric of NAS, i.e. training under a fixed policy, is biased and can mistakenly reject exemplary architectures with appropriate policy.

\begin{table*}
\caption{The clear difference between our DAAS and two closely related concurrent works.}
\label{tab:comparison_related_work}
\centering
\resizebox{.98\textwidth}{!}{
\begin{tabular}{|m{48pt}|m{62pt}|m{135pt}|m{235pt}|}
\hline
\textbf{Method} & \textbf{Target} & \textbf{Search Space} & \textbf{Search Method} \\
\hline
Combination of DARTS and Faster-AA~\cite{joint_nasaa} & AA+NAS & \textbf{1) AA:} the same search space of Faster-AA; \textbf{2) NAS:} the same search space of DARTS. & \textbf{1)} Bi-level optimization for both AA and NAS; \textbf{2)} GPU memory consuming since it simply combines DARTS and Faster-AA; \textbf{3)} Gumbel reparameterization technique to estimate gradients for augmentation parameters.  \\
\hline
DHA~\cite{dha} & AA+NAS+HPO & \textbf{1) AA:} types of augmentation policy; \textbf{2) NAS:} the same search space of DARTS; \textbf{3) HPO:} L2 regularization and learning rate.  & \textbf{1)} One-level optimization for AA and NAS task, and bi-level optimization for HPO task; \textbf{2)} Memory efficient based on sparse coding (ISTA-NAS); \textbf{3)} Gumbel reparameterization technique to estimate gradients for augmentation parameters. \\
\hline
DAAS (ours) & AA+NAS & \textbf{1) AA:} types of augmentation policy, and the application probability and magnitude for each pre-processing operator; \textbf{2) NAS:} the same search space of DARTS. & \textbf{1)} Bi-level optimization for both AA and NAS; \textbf{2)} Memory efficient NAS method by sampling a subset of supernet edges at each iteration; \textbf{3)} Policy gradient based algorithm for AA task, which reduces the computation cost of gradient estimation for augmentation parameters. \\
\hline
\end{tabular}
}
\vspace{-5pt}
\end{table*}

This paper aims to search for architecture and suitable augmentation policies jointly.
Specifically, we compare the difference between the bi-level optimization problems in NAS and AA and show the un-differentiable issue in AA task. Unlike the prior AA works~\cite{dada,faster-aa} that utilize biased gradient approximation and Gumbel reparameterization trick, this work proposes a brand-new search method based on the policy gradient algorithm that can simplify the computation of second-order partial derivatives as well.
Overall, we construct the bi-level optimization model for joint search and propose an efficient method, named DAAS.
Extensive experiments verify the effectiveness of our method. 
The contributions of this work can be summarized as follows.

\textbf{1) Policy gradient based second-order approximation for Auto-augmentation.}
This work shows that the differentiable assumption in prior methods for AA task~\cite{dada,faster-aa} is biased since they manually define the gradients w.r.t. non-differentiable augmentation parameters as one. In contrast, we refine the second-order approximation based on the policy gradient algorithm, which overcomes the above problem and reduces the computational cost by simplifying second-order partial derivatives.

\textbf{2) Differentiable joint search framework for AA and NAS.}
To our best knowledge, this is one of the pioneers (together with the arxiv works~\cite{joint_nasaa,dha}) that explore the joint search for AA and NAS in a differentiable manner. By constructing a bi-level optimization model to update the search parameters\footnote{The search parameters are made of architecture parameters for NAS and augmentation parameters for AA task.} and network weights alternately, our method can find effective combinations of architecture and augmentation policies in one GPU-day. Ablation studies are also conducted to verify the superiority of joint search against independent search for AA and NAS.

\textbf{3) Strong performance and fast speed.}
Our framework surpasses the prior NAS methods by almost 0.5\% on CIFAR-10 and 1.0\% on ImageNet, showing the necessity of combining AA with NAS tasks. In particular, the discovered architecture combined with the searched augmentation policies achieves 97.91\% accuracy on CIFAR-10 and 76.6\% top-1 accuracy on ImageNet.
Moreover, our AA algorithm is also independently evaluated and achieves competitive and even better performance than current AA methods.

\section{Related Work}
\label{sec:related_work}

\textbf{Differentiable Neural Architecture Search.}
DARTS \cite{liu2018darts} builds a cell-based supernet and introduces architecture parameters to represent the importance of operations. Though DARTS reduces the search cost to a few GPU-days, it suffers high GPU memory cost. PC-DARTS~\cite{Xu2020PC-DARTS} makes use of partial connections instead of full-fledged supernet. MergeNAS~\cite{wang2020mergenas} merges  parametric operations into one convolution. 
GDAS \cite{dong2019searching} adopts the Gumbel reparameterization technique to sample a sub-graph of the supernet at each iteration to reduce GPU memory cost. ROME~\cite{rome} reveals the instability issue in GDAS and stabilizes the search by topology disentanglement and gradient accumulation.

\textbf{Auto-augmentation.}
\citet{AA} first adopt reinforcement learning for the auto-augmentation task, but it requires searching for thousands of GPU-days. Fast-AA~\cite{fast-aa} introduces Bayesian Optimization to speed up the searching.
DADA~\cite{dada} introduces trainable augmentation parameters and adopts the Gumbel technique to update it with network weights by gradient descent algorithm alternately. Faster-AA~\cite{faster-aa} also utilizes the Gumbel technique and regards the AA task as a density matching problem. However, they both suffer a non-differentiable problem, which leads to biased gradient estimation for augmentation parameters.
In contrast, this work introduces to update augmentation parameters by policy gradient algorithm, which overcomes the non-differentiable problem with less computational cost.

\textbf{Joint searching for AutoML. }
Auto-augmentation (AA), Neural Architecture Search (NAS), and Hyper-parameter Optimization (HPO) are three popular branches of AutoML.
Recent works~\cite{autoHAS,fbnet-v3,tabularHAS,zela2018towards} have explored joint search for NAS and HPO based on reinforcement learning and performance predictor.
Our work is one of the pioneers exploring the differentiable joint search for AA and NAS. In particular, it essentially differs from the concurrent arxiv works~\cite{joint_nasaa,dha}. The first work combines Faster-AA and DARTS, which requires large GPU memory and performs worse than independent search. The second work, named DHA~\cite{dha}, regards AA and NAS as a one-level optimization problem, which oversimplifies the joint searching problem. In contrast, we construct a bi-level optimization model, which seeks better augmentation policies given the searched architecture and vice versa.
Besides, DHA only searches coarse augmentation policies and ignores the fine-grained search for application probability and magnitude for pre-processing operations. In comparison, our method searches in a more detailed search space and is still cost-effective. Table~\ref{tab:comparison_related_work} shows a clear difference between our work and the two papers. As shown in our experiments, our method also outperforms DHA under the same setting.


\section{Preliminaries and Search Space Choice}
\label{sec:preliminary}

\subsection{Preliminaries of Differentiable Methods}
\label{subsec:preliminary}
Differentiable NAS (DARTS) is first introduced in~\cite{liu2018darts}. By constructing a supernet with normal cells and reduction cells, it introduces architecture parameters $\bm{\alpha}$ to represent the importance of candidate operations and connections and regards NAS as a bi-level optimization problem. 
To reduce the GPU memory and alleviate the topology gap between the supernet and final network, recent works~\cite{dong2019searching,xie2018snas,rome} utilize the Gumbel technique to sample and activate a subset of operations. The bi-level optimization model is constructed as Eq.~\ref{eq:nas_bi_level}, where $\bm{\theta}_z^*$ is the optimal operation weights for the sampled architecture $z$, and $L_v, L_t$ denotes the loss function in the validation and training procedure.
\begin{align}
\min_{\bm{\alpha}} \quad  &L_v(\bm{\theta}_z^*, z; D_{val})  \label{eq:nas_bi_level} \\
\mathrm{s.t.} \quad &z\sim p(z;\bm{\alpha}), \quad \bm{\theta}_z^* = \arg\min_{\bm{\theta}} L_t(\bm{\theta}, z; D_{train}) \nonumber
\end{align}
Inspired by DARTS~\cite{liu2018darts} and GDAS~\cite{dong2019searching}, DADA~\cite{dada} adopts the differentiable based method and Gumbel technique for auto-augmentation. Specifically, it introduces augmentation parameters $\bm{\gamma}$ to represent the importance and hyper-parameters for augmentation policies and formulates the bi-level optimization model as Eq.~\ref{eq:aa_bi_level}, where $\Gamma$ denotes the sampled augmentation policy, and $D_{train}, D_{val}$ denote the training set and validation set.
\begin{align}
		\min_{\bm{\gamma}} \quad  &L_v(\bm{\theta}^*; D_{val}) \label{eq:aa_bi_level}\\
		\mathrm{s.t.} \quad &\Gamma\sim p(\Gamma;\bm{\gamma}), \quad \bm{\theta}^* =  \arg\min_{\bm{\theta}} L_t(\bm{\theta}; \Gamma(D_{train})) \nonumber
\end{align}

\subsection{Search Spaces}
\label{subsec:search_space}
In line with the mainstream of existing works, the joint search space is a Cartesian product of augmentation policy and architecture spaces. Specifically, we refer to the DADA~\cite{dada} and 
pair up image pre-processing operations and build various candidate policies. For the NAS task, we adopt the search space of DARTS~\cite{liu2018darts} and construct a supernet containing all the candidate operations and connections. 

\textbf{Data Augmentation Search Space.}
Similar to DADA~\cite{dada} and Fast-AA~\cite{fast-aa}, we construct a set of image pre-processing operations $\mathcal{O}_{aug}$ with 15 candidates, including rotation, translation, and etc. Each policy $\Gamma$ contains $K$ image pre-processing operations: $\Gamma = \{o_k\}_{k=1}^K$, where $o_k\in \mathcal{O}_{aug}$. So there are total $|\mathcal{O}_{aug}|^K$ policies in our search space.
The augmentation parameters $\bm{\gamma}$ contain three parts: the sampling weights $\bm{\pi}$, the operation probability $\bm{p}$, and the operation magnitude $\bm{\delta}$.
Specifically, we define sampling weights $\bm{\pi}\in\mathbb{R}^{K\times|\mathcal{O}_{aug}|}$ to represent the sampling probability for each pre-processing operation. Therefore, the $k$-th operation can be sampled as: $o_k\sim \tilde{\bm{\pi}}(o;\bm{\pi}_k)=\mathrm{softmax}(\bm{\pi}_k)\in\mathbb{R}^{|\mathcal{O}_{aug}|}$.
Besides, for $i$-th policy, each pre-processing operation $o^i_k$ has two parameters: the probability $p^i_k$ and magnitude $m^i_k$ to apply that operation. Referring to AA~\cite{AA}, we discretize the range of magnitude into 10 values (uniform spacing) and use $\bm{\delta}^i_k\in \mathbb{R}^{10}$ to represent the importance of each magnitude candidate, so that each candidate magnitude will be selected with probability $m^i_k \sim p(m^i_k;\bm{\delta}^i_k) = \text{softmax}(\bm{\delta}^i_k)$.
Overall, the probability of a specific policy $\Gamma^{i}=\{ o^{i}_k, y_k^{i}, p_k^i, m_k^{i} \}_{k=1}^K$ can be formulated as Eq.~\ref{eq:prob_policy}, where $y_k^i\in \{0,1\}$ indicates whether to apply the operation $o_k^i$, and $m_k^i$ is the sampled magnitude.
\begin{equation}
    p(\Gamma^{i}) = \prod \limits_{k=1}^{K} p(o^{i}_k;\bm{\pi}_k) \cdot (1-p_k^i)^{(1-y_k^i)} \cdot \left[p_k^i \cdot p(m_k^i; \bm{\delta}_k^i) \right]^{y_k^i} 
    \label{eq:prob_policy}
\end{equation}

\textbf{Architecture Search Space.}
A supernet is stacked by normal and reduction cells. Each cell contains $N$ nodes $\{x_i\}_{i=1}^{N}$ representing latent feature maps. The outputs of all intermediate nodes are concatenated as the output of the cell. There is an parallel edge $\bm{e}_{i,j}$ between every two nodes $x_i, x_j$, which integrates all the candidate operations, i.e. $\bm{e}_{i,j} = \{ o_{i,j}| o\in \mathcal{O}_{arch}\}$, where $\mathcal{O}_{arch}$ is the candidate operation set, including separable convolutions, pooling, and identity operations.
We define architecture parameters $\bm{\beta}$ and $\bm{\alpha}$ to represent the importance of edges and operations in the supernet, respectively.
At each iteration, we sample two parallel edges for each node based on $\bm{\beta}$ and sample one operation for each parallel edge based on $\bm{\alpha}$.
The output of node $x_j$ can be computed as follows:
\begin{align}
&x_{j}=\sum_{i<j} \bigg[ B_{i,j}\cdot \sum_{o\in\mathcal{O}_{arch}} A_{i,j}^o o_{i,j}(x_{i}) \bigg] \\
&\mathrm{s.t.} \quad A_{i,j}^o, B_{i,j}\in \{0, 1\}, \quad \sum_{i<j}B_{i,j} = 2, \sum_{o\in\mathcal{O}_{arch}}A_{i,j}^o = 1 \nonumber
\end{align}
where $\bm{B}_{:,j}=[B_{i,j}]_{i<j}$ is a two-hot vector denoting the sampled edges and $\bm{A}_{i,j}=[A^o_{i,j}]_{o\in\mathcal{O}_{arch}}$ is a one-hot vector denoting the sampled operation for the parallel edge $\bm{e}_{i,j}$.

\section{DAAS Method}
\label{method}
We first construct the bi-level optimization model for joint search in Sec.~\ref{subsec:bi_level} and then detail the search method in Sec.~\ref{subsec:joint_aa}-\ref{subsec:joint_weight}. Next, we introduce the strategy to derive the final augmentation policy and architecture in Sec.~\ref{subsec:derive}. Finally, we analyze the superiority of our method in Sec.~\ref{subsec:discussion}.

\subsection{Bi-level Optimization for Joint Search}
\label{subsec:bi_level}
Bi-level optimization for joint search can be formulated as:
\begin{align}
		\min_{\bm{\alpha},\bm{\beta},\bm{\gamma}} \quad  &\bar{L}_v = \mathbb{E}_{z\sim p(z;\bm{\alpha},\bm{\beta})}\left[L_v(\bm{\theta}_z^*(\bm{\gamma}), z; D_{val}) \right]  \label{eq:joint_bi_level} \\
		\mathrm{s.t.} \quad   &\bm{\theta}_z^*(\gamma) = \arg\min_{\bm{\theta}}
		\mathbb{E}_{\Gamma\sim p(\Gamma;\bm{\gamma})}\left[L_t(\bm{\theta}, z; \Gamma(D_{train}))\right] \nonumber
\end{align}
where $\bm{\alpha}$ and $\bm{\beta}$ represent the importance of candidate operations and connections, $\bm{\gamma}$ represent the importance of candidate augmentation policies and the corresponding hyper-parameters, and $\bm{\theta}$ is the network weights in the supernet. 
Our bi-level optimization is more than simply combining the optimization model of NAS and AA.

On one hand, we refine the optimization target by the expectation of $L_v$ w.r.t. $p(z)$ and $p(\Gamma)$.
different from the optimization in GDAS (Eq.~\ref{eq:nas_bi_level}) and DADA (Eq.~\ref{eq:aa_bi_level}) that directly regarding $L_v$ of one sampled candidate as the optimization target, we argue that the goal of bi-level optimization in Eq.~\ref{eq:joint_bi_level} is to obtain a proper sampling probabilities for candidate augmentation policies and architectures, as represented by the augmentation parameters $\bm{\gamma}$ and architecture parameters $\bm{\alpha}, \bm{\beta}$. Consequently, the expectation of loss $\bar{L}$ in Eq.~\ref{eq:joint_bi_level} over sampling probability should be the optimization target.


On the other hand, we argue that estimating gradient w.r.t. augmentation parameters $\bm{\gamma}$ by the chain rule as DADA~\cite{dada} is impracticable since the optimization target $L_v$ can be non-differentiable w.r.t. $\bm{\gamma}$.
Specifically, since data augmentation is not applied on validate data, the gradient $\nabla_{\bm{\gamma}}L_v = \nabla_{\bm{\theta}^*}L_v\cdot\nabla_{\bm{\gamma}}\bm{\theta}^*$ only contains the second-order derivative item according to the chain rule.
DADA adopts the second-order approximation by estimating $\bm{\theta}^*\approx\bm{\theta}-\eta\nabla_{\bm{\theta}}L_t$, where $\eta$ is the learning rate to train operation weights. So that $\nabla_{\bm{\gamma}}L_v\approx
     \nabla_{\bm{\theta}^*}L_v\cdot(-\eta)\nabla^2_{\Gamma,\bm{\theta}}L_t\cdot \nabla_{\bm{\gamma}}\Gamma$.
However, the policy $\Gamma$ can be non-differentiable w.r.t. $\bm{\gamma}=\{\bm{\pi}, \bm{p}, \bm{m}\}$, e.g., the gradient w.r.t. operation magnitude $\nabla_{\bm{m}}\Gamma$ is nonexistent, making $\nabla_{\bm{\gamma}}L_v$ nonexistent. To this end, DADA manually defines $\nabla_{\bm{m}}\Gamma \equiv 1$, which is groundless and incorrect. 
In contrast, we refine the optimization target for $\bm{\theta}^*$ as the expectation of $L_t$, making it possible to directly estimate $\nabla_{\bm{\gamma}}\bm{\theta}^*$ by policy gradient algorithm with no need to define a nonexistent gradient (detailed in Sec.~\ref{subsec:joint_aa}).

This work attempts to solve the above optimization problem (in Eq.~\ref{eq:joint_bi_level}) by a differentiable method. 
To train augmentation parameters $\bm{\gamma}$, we first obtain the gradients w.r.t. $\bm{\gamma}$ by the chain rule: $\nabla_{\bm{\gamma}}\bar{L}_v = \nabla_{\bm{\theta}^*}\bar{L}_v\cdot\nabla_{\bm{\gamma}}\bm{\theta}^*$, and then estimate $\nabla_{\bm{\gamma}}\bm{\theta}^*$ by policy gradient algorithm. 
To update architecture parameters, we utilize the first-order approximation~\cite{liu2018darts,dong2019searching,rome} and estimate gradients w.r.t. $\bm{\alpha}$ as $\nabla_z \bar{L}_v\cdot \nabla_{\bm{\alpha}}z$. 
We detailed the computation of gradients in the following.

\subsection{Policy Gradients for Augmentation Parameters $\bm{\gamma}$}
\label{subsec:joint_aa}

Given an architecture $z$, we apply one-step SGD to estimate the optimal network weights $\bm{\theta}^*$ in Eq.~\ref{eq:joint_bi_level} as follows:
\begin{equation}
\bm{\theta}^*\approx \bm{\theta}-\eta\nabla_{\bm{\theta}}\mathbb{E}_{\Gamma\sim p(\Gamma;\bm{\gamma})}\left[L_t(\bm{\theta}, z; \Gamma(D_{train}))\right] \triangleq \bm{\theta}'
\label{eq:optimal_theta}
\end{equation}
where $\eta$ is the learning rate, and $\Gamma$ is the augmentation policy. 
Therefore, the gradient of search loss w.r.t. augmentation parameters $\bm{\gamma}$ can be formulated as follows:
\begin{align}
    \nabla_{\bm{\gamma}}\bar{L}_v = \mathbb{E}_z \bigg\{ &\nabla_{\bm{\theta}'} L_v(\bm{\theta}', z; D_{val}) \times (-\eta)   \label{eq:gradient_policy_v0}\\
    &\times \nabla^2_{\bm{\theta},\bm{\gamma}}\mathbb{E}_{\Gamma\sim p(\Gamma;\bm{\gamma})}\big[L_t(\bm{\theta}, z; \Gamma(D_{train}))\big]
    \bigg\} \nonumber
\end{align}

Though $\nabla_{\bm{\gamma}}\Gamma$ is nonexistent, the sampling probability $p(\Gamma)$ in Eq.~\ref{eq:prob_policy} is differentiable w.r.t. $\bm{\gamma}$. Therefore, we adopt policy gradient algorithm.
Specifically, the expectation of $L_t$ in Eq.~\ref{eq:gradient_policy_v0}, $\mathbb{E}_{\Gamma\sim p(\Gamma;\bm{\gamma})}\left[L_t(\bm{\theta}, z; \Gamma(D_{train}))\right]$, can be formulated as $\sum\left[L_t(\bm{\theta}, z; \Gamma(D_{train})) \cdot p(\Gamma)\right]$, then the second-order partial derivative can be simplified as: 
\begin{align}
    &\nabla^2_{\bm{\theta},\bm{\gamma}}\mathbb{E}_{\Gamma\sim p(\Gamma;\bm{\gamma})}\big[L_t(\bm{\theta}, z; \Gamma(D))\big]  \label{eq:second_order_partial} \\
    =& \nabla^2_{\bm{\theta},\bm{\gamma}} \sum\big[L_t(\bm{\theta}, z; D') \cdot p(\Gamma)\big] \nonumber \\
    =& \sum \big[ \nabla_{\bm{\theta}}L_t(\bm{\theta}, z; D') \cdot \nabla_{\bm{\gamma}}\log p(\Gamma) \cdot p(\Gamma) \big] \nonumber \\
    =& \mathbb{E}_{\Gamma\sim p(\Gamma;\bm{\gamma})}\big[\nabla_{\bm{\theta}}L_t(\bm{\theta}, z; D')\cdot \nabla_{\bm{\gamma}}\log p(\Gamma)\big] \nonumber
\end{align}
where $D'=\Gamma(D)$ is the augmented data, and $p(\Gamma)$ is the probability of a specific policy $\Gamma$ as formulated as Eq.~\ref{eq:prob_policy}.
Eq.~\ref{eq:second_order_partial} simplifies the intractable second-order partial derivative by a vector multiplication of two first-order gradients.
Consequently, the gradient $\nabla_{\bm{\gamma}}\bar{L}_v$  can be computed as Eq.~\ref{eq:gradient_policy_v1}, where $\bm{\theta}'$ is defined in Eq.~\ref{eq:optimal_theta}.
\begin{align}
    \nabla_{\bm{\gamma}}\bar{L}_v &= -\eta \mathbb{E}_z \bigg\{
        \nabla_{\bm{\theta}'}L_v(\bm{\theta}', z; D_{val})  \label{eq:gradient_policy_v1}  \\
    &\quad \times \mathbb{E}_{\Gamma\sim p(\Gamma;\bm{\gamma})}\big[\nabla_{\bm{\theta}}L_t(\bm{\theta}, z; D')\cdot \nabla_{\bm{\gamma}}\log p(\Gamma)\big]
    \bigg\} \nonumber \\
    &\approx \frac{1}{NM}\sum_{i=1}^{N}  \bigg\{\nabla_{\bm{\theta}'}L_v(\bm{\theta}', z^{(i)}; D_{val}) \nonumber \\ &\quad \times \sum_{j=1}^M \left[\nabla_{\bm{\theta}}L(\bm{\theta}, z^{(i)}, D')\cdot \nabla_{\bm{\gamma}}\log p(\Gamma^{(i,j)})\right] \bigg\}\nonumber
\end{align}
$N, M$ are the sampling numbers for architectures and policies, respectively. We set $N$=5 and $M$=2 in our experiment.

\subsection{Gradients for Architecture Parameters $\bm{\alpha}, \bm{\beta}$}
\label{subsec:joint_nas}
We adopt Gumbel reparameterization technique to estimate the gradients w.r.t. architecture parameters. Suppose  $\tilde{\bm{\alpha}}_{i,j}^o = \frac{\exp(\alpha_{i,j}^o)}{\sum_{o'\in\mathcal{O}}\exp(\alpha_{i,j}^{o'})}$ and $\tilde{\bm{\beta}}_{i,j} = \frac{\exp(\beta{i,j})}{\sum_{k<j}\exp(\beta_{k,j})}$ are normalized architecture parameters. The sampled operation on edge $e_{i,j}$ can be represented by a one-hot vector $\bm{A}_{i,j}$ as Eq.~\ref{eq:gumbel_nas_alpha}, where $g_{i,j}^o$ are sampled from Gumbel(0,1) distribution.
\begin{align}
    \tilde{\bm{A}}_{i,j}^{o} &= \frac{\exp\left[(\log\tilde{\bm{\alpha}}_{i,j}^{o}+\bm{g}_{i,j}^{o})/\tau\right]}   {\sum_{o'=1}^{|\mathcal{O}|} \exp\left[(\log\tilde{\bm{\alpha}}_{i,j}^{o'}+\bm{g}_{i,j}^{o'})/\tau\right]} \nonumber \\
    \bm{A}_{i, j} &= \mathrm{one}\underline{\hbox to 0.2cm{}} \mathrm{hot} \left[\arg\max_{o\in\mathcal{O}}{\tilde{\bm{A}}}_{i,j}^{o}\right]
    \label{eq:gumbel_nas_alpha}
\end{align}
Similarly, a two-hot vector $\bm{B}_{:,j}=[B_{i,j}]_{i<j}$ denotes the sampled edges for node $j$, whose item is shown as Eq.~\ref{eq:gumbel_nas_beta}, where $h_{i,j}$ are sampled from Gumbel(0,1) distribution.
\begin{align}
    \tilde{\bm{B}}_{i,j} &= \frac{\exp\left((\log\tilde{\bm{\beta}}_{i,j}+\bm{g}_{i,j})/\tau\right)}   {\sum_{i'<j} \exp\left((\log\tilde{\bm{\beta}}_{i',j}+\bm{g}_{i',j})/\tau\right)} \nonumber \\
    \bm{B}_{i,j} &= \left\{
        \begin{aligned}
        1, &\quad i \in \arg\mathop{\mathrm{top2}}\limits_{i'<j} (\tilde{\bm{B}}_{i',j}) \\
        0, &\quad otherwise
        \end{aligned}.
    \right.
    \label{eq:gumbel_nas_beta}
\end{align}
The computational cost and GPU memory requirement can be significantly reduced since only the sampled edges and operations are activated in the forward pass. 

A sampled architecture $z$ can be determined as: $z=\{\bm{B}_{i,j} | 2\leq j \leq N, 1\leq i < j\} \cup \{\bm{A}_{i,j}^o | 2\leq j \leq N, 1\leq i < j, o\in\mathcal{O}_{arch}\}$. We can then estimate the gradients $\nabla_{\bm{\alpha}}z=\nabla_{\bm{\alpha}}\tilde{\bm{A}}$ and $\nabla_{\bm{\beta}}z=\nabla_{\bm{\beta}}\tilde{\bm{B}}$ based on Gumbel-softmax technique~\cite{wu2018fbnet,dong2019searching,rome}. 
Therefore, the gradient of search loss $\bar{L}_v$ w.r.t. architecture parameters $\bm{a}=\{\bm{\alpha}, \bm{\beta}\}$ can be formulated as Eq.~\ref{eq:gradient_arch}, where $D_{val}$ is a batch of data.
\begin{align}
    \nabla_{\bm{a}}\bar{L}_v &= \mathbb{E}_{z\sim p(z;\bm{\alpha},\bm{\beta})} \left[ \nabla_z L_v(\bm{\theta}, z; D_{val}) \cdot \nabla_{\bm{a}} z \right]   \label{eq:gradient_arch} \\
    &\approx \frac{1}{N}\sum_{i=1}^{N}\nabla_z L_v(\bm{\theta}, z^{(i)}; D_{val}) \cdot \nabla_{\bm{a}} z^{(i)} \nonumber
\end{align}

\begin{algorithm}[tb!]
	\caption{DAAS: Differentiable Architecture and Augmentation Policy Search}
	\label{alg:DAAS}
	\begin{algorithmic}
\STATE {\bfseries Parameters of supernet:} Initialized operation weights $\bm{\theta}$; Architecture parameters $\bm{\alpha}$, $\bm{\beta}$; Augmentation parameters $\bm{\gamma}$; Learning rates $\xi_{\bm{\alpha}}$, $\xi_{\bm{\beta}}$, $\xi_{\bm{\gamma}}$, and $\eta_{\bm{\theta}}$.
    \REPEAT
\STATE		\textbf{1) }Sample batches of data $D_{train}$, $D_{val}$, candidate architectures $\{z^{(i)}\}$, and policies $\{\Gamma^{(i,j)}\}$;
		
\STATE			\textbf{2) }Estimate the gradients for $\bm{\alpha}, \bm{\beta}$ by Eq.~\ref{eq:gradient_arch} and $\bm{\gamma}$ by Eq.~\ref{eq:gradient_policy_v1}, and update by gradient descent:
		$\bm{a}\leftarrow \bm{a}-\xi_{\bm{a}}\nabla_{\bm{a}}\bar{L}_v, \bm{a}\in\{\bm{\alpha},\bm{\beta},\bm{\gamma}\}$;

\STATE	    \textbf{3) }Estimate gradients for supernet weights by Eq.~\ref{eq:gradient_theta} and update by gradient descent: $\bm{\theta}\leftarrow \bm{\theta}-\eta_{\bm{\theta}}\nabla_{\bm{\theta}}\bar{L}_t$;
   \UNTIL{Converged}
    \end{algorithmic}
\end{algorithm}

\subsection{Training for Supernet Weights $\bm{\theta}$}
\label{subsec:joint_weight}
Weights for candidate architecture $z$ are directly obtained from the supernet, so we should train supernet weights $\bm{\theta}$ to adapt various architectures. 
Moreover, given a specific architecture $z$,  the optimization target to train network weights $\bm{\theta}_z$ should be the expectation $\mathbb{E}_{\Gamma}\left[L_t(\bm{\theta},z;\Gamma(D_{train}))\right]$ according to the bi-level optimization in Eq.~\ref{eq:joint_bi_level}. Overall, the target to train $\bm{\theta}$ can be formulated as follows.
\begin{equation}
    \bm{\theta}^* = \arg\min_{\bm{\theta}} \mathbb{E}_{z\sim p(z)} \bigg\{ \mathbb{E}_{\Gamma\sim p(\Gamma;\bm{\gamma})}\big[L_t(\bm{\theta}, z; \Gamma(D_{train}))\big]
    \bigg\} \nonumber
\end{equation}
Consequently, the gradients for supernet weights $\bm{\theta}$ can be computed as Eq.~\ref{eq:gradient_theta}, which can be estimated by the mean of multiple samples.
\begin{align}
    \nabla_{\bm{\theta}}\bar{L}_t &=  \mathbb{E}_{z\sim p(z)} \bigg\{ \mathbb{E}_{\Gamma\sim p(\Gamma;\bm{\gamma})}\big[\nabla_{\bm{\theta}}L_t(\bm{\theta}, z; \Gamma(D_{train}))\big]
    \bigg\} \nonumber \\
    &\approx \frac{1}{NM}\sum_{i=1}^N \sum_{j=1}^M \nabla_{\bm{\theta}}L_t(\bm{\theta}, z^{(i)}, \Gamma^{(i,j)}(D_{train})) \label{eq:gradient_theta}
\end{align}
where $N, M$ are the sampling numbers.
The algorithm of our DAAS is outlined in Alg.~\ref{alg:DAAS}.

\subsection{Deriving Policies and Architecture}
\label{subsec:derive}
The final architecture is derived according to the magnitude of architecture parameters. Specifically, we preserve two edges for each node (based on $\bm{\beta}$) and one operation on each selected edge (based on $\bm{\alpha}$).
As for the augmentation policies, we pair up the image pre-processing operations and enumerate all policies and its sampling probability: for policy $\Gamma^{i} = \{o_k^{i}\}_{k=1}^K$, its sampling probability is $\tilde{\bm{\pi}}(\Gamma^i) = \prod_{k=1}^K \tilde{\bm{\pi}}(o_k^i;\bm{\pi}_k)$. 
For each operation $o_k^i$ in the policy $\Gamma^i$, its application probability is $p_k^i$ and its application magnitude is $m_k^i = \arg\max p(m_k^i; \bm{\delta}_k^i)$.

The derived architecture is evaluated by training from scratch with the discovered augmentation policies.
At each iteration, we sample one policy $\Gamma=\{o_k\}_{k=1}^K$ based on the sampling probability $\tilde{\bm{\pi}}(\Gamma)$ and augment the input data by the K pre-processing operations in sequence.

\begin{table*}[tb!]
	\caption{Evaluation on CIFAR-10 (left) and CIFAR-100 (right). We report the best (in the first block) and average (in the second block) performance over four parallel tests by searching under different random seeds. $^\star$: Results are obtained by training the best architecture multiple times rather than searching for multiple times. $^\dagger$: Results of joint search for AA and NAS reported in DHA~\cite{dha}. }
		\label{tab:comparision_cifar}
	\vspace{-15pt}
\setlength{\tabcolsep}{2pt}
	\begin{center}
		\begin{scriptsize}
	    \begin{minipage}[t]{0.48\textwidth}
	 	\center
	 	\resizebox{\textwidth}{!}{
			\begin{tabular}{lcHcc} 	
				\toprule		
			\multirow{2}{*}{\textbf{CIFAR-10}} 	  &  \textbf{\scriptsize{Params}}  &  \textbf{\scriptsize{FLOPs}}  &  \textbf{Error}  &  \textbf{Cost}  \\
				 & \scriptsize{(M)}  & \scriptsize{(M)}  & \scriptsize{(\%)} & \scriptsize{GPU Days}   \\
				\midrule
				NASNet-A  \citeyearpar{zoph2017learning}   &  3.3  &  608$^\dagger$   &   2.65  &  2000  \\
				ENAS \citeyearpar{pham2018efficient}  &  4.6  &  626$^\dagger$ &  2.89  & 0.5    \\	

				DARTS \citeyearpar{liu2018darts}  &  3.3  &  528$^\dagger$  &  3.00$\pm0.14^\star$  &  0.4  \\ 
			    P-DARTS \citeyearpar{chen2019progressive}  &  3.4  & 
			    532$^\dagger$  &  2.50  &  0.3 \\
				SNAS \citeyearpar{xie2018snas}  &  2.8  &  422$^\dagger$  &  2.85$\pm0.02^\star$  &  1.5\\
				GDAS \citeyearpar{dong2019searching}  &  3.4  &  519$^\dagger$  &  2.93  & 0.2\\
				PC-DARTS \citeyearpar{Xu2020PC-DARTS}  &  3.6  &  558$^\dagger$ &  2.57  &  0.1  \\ 
			    DARTS- \citeyearpar{darts-} & 3.5  & 568  &  2.50  & 0.4\\
				\textbf{DAAS} (best) & 4.4 &  & \textbf{2.09} & 1.0 \\
				\midrule
				R-DARTS \citeyearpar{zela2020understanding}  & - &  - &  2.95$\pm$0.21 & 1.6 \\
				SDARTS-ADV \citeyearpar{chen2020stabilizing}  & 3.3  & -  &  2.61$\pm$0.02  & 1.3\\
				ROME~\citeyearpar{rome} & 3.7 & & 2.58$\pm$0.07 & 0.3 \\
				DARTS-~\citeyearpar{darts-} & 3.5  & 583$\pm$22  &  2.59$\pm$0.08  & 0.4\\ 
			    DARTS+Faster-AA~\citeyearpar{joint_nasaa}& - & &2.60$\pm$0.03 & - \\
			    DHA$^\dagger$(AA+NAS)~\citeyearpar{dha} & - & & \textbf{2.22$\pm$0.13} & 2.7 \\
				\textbf{DAAS} (avg.) & 4.0 & & 2.24$\pm$0.10 & 1.0 \\
				\bottomrule
			\end{tabular}
		}
		\end{minipage}
		\begin{minipage}[t]{0.48\textwidth}
        \center
        \resizebox{\textwidth}{!}{
			\begin{tabular}{lcHcc} 	
				\toprule		
			\multirow{2}{*}{\textbf{CIFAR-100}} 	  &  \textbf{Params}  &  \textbf{FLOPs} &  \textbf{Error } &  \textbf{Cost}  \\
				 & \scriptsize{(M)}  & \scriptsize{(M)}  & \scriptsize{(\%)} & \scriptsize{GPU Days}   \\
				\midrule
				AmoebaNet \citeyearpar{real2019regularized} & 3.1 & &  18.93 & 3150  \\
				PNAS \citeyearpar{liu2018progressive} & 3.2 & & 19.53 & 150 \\
				ENAS \citeyearpar{pham2018efficient}  &  4.6  &    &  19.43  & 0.45    \\
				DARTS \citeyearpar{liu2018darts}  & - & & 20.58$\pm$0.44$^\star$ &  0.4  \\ 
				P-DARTS \citeyearpar{chen2019progressive}  &  3.6  &   &  17.49  &  0.3 \\
				GDAS \citeyearpar{dong2019searching}  &  3.4  &    &  18.38  & 0.2 \\
				ROME~\citeyearpar{rome} & 4.4 & & 17.33 & 0.3 \\
				DARTS- \citeyearpar{darts-} &  3.4 &   & 17.16  & 0.4\\
				\textbf{DAAS} (best) & 3.7 & & \textbf{15.20} & 1.0 \\
				\midrule
				R-DARTS \citeyearpar{zela2020understanding}  & - &  - &  18.01$\pm$0.26 & 1.6 \\
				ROME~\citeyearpar{rome} & 4.4 & & 17.41$\pm$0.12 & 0.3 \\
				DARTS-~\citeyearpar{darts-} & 3.3& & 17.51$\pm$0.25 & 0.4 \\
				DARTS+Faster-AA~\citeyearpar{joint_nasaa}& - & &16.19$\pm$0.49 & -\\
			    DHA$^\dagger$(AA+NAS)~\citeyearpar{dha} & - & & 16.45$\pm$0.03 & 2.7 \\
				\textbf{DAAS} (avg.) & 3.8 & & \textbf{15.37$\pm$0.31} & 1.0 \\
				\bottomrule
			\end{tabular}
        }
		\end{minipage}
		\end{scriptsize}
	\end{center}
	\vspace{-15pt}
\end{table*}

\subsection{Discussion}
\label{subsec:discussion}
\textbf{Necessity to refine the optimization target.}
In many prior differentiable based works for AA~\cite{dada,faster-aa} and NAS~\cite{dong2019searching,xie2018snas}, only one sampling is considered to compute the search loss. Rethinking the optimization target, we aim to obtain the proper probability $p(\Gamma;\bm{\gamma})$ and $p(z;\bm{\alpha}, \bm{\beta})$ to sample policy and architecture.
Therefore, a single sample is insufficient to represent the distribution, resulting in biased gradient estimation for architecture and augmentation parameters. Recent works~\cite{ddas,rome} also point out the importance of multiple sampling for AA or NAS.
This work constructs a bi-level optimization model for joint search in Eq.~\ref{eq:joint_bi_level}.

\textbf{Strength of policy gradient algorithm for AA.}
Recent differentiable based AA works~\cite{dada,faster-aa} adopt Gumbel reparameterization technique to estimate the gradient of loss w.r.t. the augmentation parameters. However, such estimation is biased due to \emph{manually defined gradients for magnitude parameters}. The augmentation operation $o(D;p,m)$ is non-differentiable w.r.t. the magnitude $m$, so they have to manually define $\nabla_{m}o(D;p,m) \equiv 1$ to satisfy the chain rule for gradient estimation.
In this work, we adopt a policy gradient algorithm to estimate the gradient for $\bm{\gamma}$ after refining the bi-level optimization target for AA in Eq.~\ref{eq:joint_bi_level}.
With no need for manually defined gradients for magnitude parameters, our method simplifies the second-order partial derivative as a multiplication of two first-order derivatives (Eq.~\ref{eq:second_order_partial}), which is more efficient than the Gumbel-softmax reparameterization technique.

\textbf{Strength of joint searching.}
On one hand, augmentation policies are coupled with architectures. Recent works on AA~\cite{dada,faster-aa,AA} search augmentation policy for different network architectures, including Wide-ResNet~\cite{wideResNet}, ShakeShake~\cite{shakeshake}, and Pyramid~\cite{pyramid}, and the discovered optimal policies differs over architectures. 
On the other hand, architectures are also related to augmentation policies. CNN can be regarded as a feature extractor and is sensitive to specific data distributions that can be affected by augmentation policies.
Joint searching for NAS and AA considers the coupling between architecture and augmentation policy and can find optimal combinations.
Results in Sec.~\ref{subsec:ablation_study} verify our analysis.

\section{Experiments}
\label{sec:experiments}
\textbf{Search Settings.}
We follow DARTS~\cite{liu2018darts} and construct a supernet by stacking 8 cells with 16 initial channels. Each cell contains $N=6$ nodes, two of which are input nodes. Two reduction cells are located at $1/3$ and $2/3$ of the total depth of the supernet.
In the search stage, we first warmup the supernet by alternately updating operation weights $\bm{\theta}$ and architecture parameters ($\bm{\alpha}$ and $\bm{\beta}$) for $20$ epochs, and then jointly search for architectures and policies for another $30$ epochs. 
We set the sampling number $N=5$ and $M=2$ by default.
For operation weights, we use SGD optimizer with 0.9 momentum; For architecture and policy weights, we adopt Adam optimizer with $\beta=(0.5, 0.999)$.
To search on CIFAR-10 and CIFAR-100, we split the training set into two parts as $D_{train}$ and $D_{val}$ to train supernet weights $\bm{\theta}$ and search parameters $\bm{\alpha},\bm{\beta},\bm{\gamma}$ respectively. 
Moreover, due to the high efficiency of our method, we directly search on ImageNet. We follow DADA~\cite{dada} and construct a surrogate dataset by selecting 120 classes.

\textbf{Evaluation Settings.}
We use standard evaluation settings as DARTS \cite{liu2018darts} by training the inferred model for 600 epochs using SGD with a batch size of 96 for CIFAR-10 and CIFAR-100.
The searched architecture is also transferred to ImageNet by stacking 14 cells with 48 initial channels.
The transferred and searched models on ImageNet are trained for 250 epochs by SGD with a batch size of 1024. In the evaluation stage, we preserve all possible policy strategies and sample one policy based on the learned sampling parameter $\bm{\pi}$ at each iteration. Note that both the search and evaluation experiments are conducted on NVIDIA V100.

\subsection{Performance Evaluation}
In this section, we first report the performance of our method on CIFAR datasets and ImageNet. Then, we show the superiority of the searched policy against the default policies used by prior NAS works. Four parallel tests are conducted on each benchmark by searching for NAS and AA under different random seeds.

\textbf{Performance on CIFAR datasets.}
Table~\ref{tab:comparision_cifar} shows the best and averaged performance over four parallel tests by searching under different random seeds. 
Compared with prior NAS works, our method achieves 97.91\% accuracy on CIFAR-10 and 84.80\% accuracy on CIFAR-100, surpassing DARTS~\cite{liu2018darts} by nearly 1\%, as shown in Table~\ref{tab:comparision_cifar}. 
Additionally, the average performance is also reported. Our discovered architectures achieve state-of-the-art on both CIFAR-10 and CIFAR-100 datasets, showing that our joint search method can stably improve the performance of NAS.
Moreover, our method can discover effective architectures and policies in 1 GPU-days, showing the high efficiency of our joint searching algorithm.

\begin{table}
\caption{Comparison on ImageNet. The top block reports the performance of models transferred from CIFAR (TF), and the bottom reports that of models directly searched on ImageNet (DS).}
\label{tab:compare_imagenet}
\centering
\resizebox{.98\columnwidth}{!}{
	\setlength{\tabcolsep}{2.0pt}
	\begin{tabular}{lcccccc} 			
	\toprule
	\multirow{2}{*}{\textbf{Models}} & \textbf{FLOPs}  & \textbf{Params} & \textbf{Top-1} & \textbf{Cost} & \multirow{2}{*}{\textbf{Way}}\\
	& \small{(M)} & \small{(M)} & \small{(\%)}  & \small{GPU days} & \\
	\midrule

	AmoebaNet-A \citeyearpar{real2019regularized} & 555  & 5.1 & 74.5& 3150&TF \\


	NASNet-A \citeyearpar{zoph2017learning}  & 564 & 5.3 &74.0 & 2000&TF \\
	PNAS \citeyearpar{liu2018progressive} & 588 & 5.1 & 74.2  &225&TF \\ 

	DARTS \citeyearpar{liu2018darts} & 574 & 4.7 & 73.3 & 0.4&TF \\
	P-DARTS \citeyearpar{chen2019progressive}& 577 & 5.1 & 75.3 & 0.3& TF \\
	FairDARTS-B \citeyearpar{chu2019fair}& 541 & 4.8 &75.1 & 0.4&TF \\
	SNAS \citeyearpar{xie2018snas}  &522&4.3&72.7 & 1.5&TF \\

	PC-DARTS  \citeyearpar{Xu2020PC-DARTS} & 586 & 5.3 & 74.9 & 0.1&TF \\ 
	GDAS \citeyearpar{dong2019searching}  &581&5.3&74.0  & 0.2&TF  \\
	ROME~\citeyearpar{rome} &576&5.2&75.3&0.3&TF \\
	\textbf{DAAS (ours)} & 698 & 6.1 & \textbf{76.6} & 1.0 & TF  \\
    \midrule
    PC-DARTS \citeyearpar{Xu2020PC-DARTS}$^\ddagger$  & 597 & 5.3 & 75.4 & 3.8&DS\\
	GDAS~\citeyearpar{dong2019searching}&405&3.6&72.5&0.8&DS \\
	ROME~\citeyearpar{rome} & 556 & 5.1 & 75.5 & 0.5&DS \\
	\textbf{DAAS (ours)} & 661 & 5.9 & \textbf{76.5} & 1.8 & DS  \\
	\bottomrule
\end{tabular}
}
\vspace{-10pt}
\end{table}

\textbf{Performance on ImageNet.}
We conduct experiments on ImageNet to verify the effectiveness of our method. We follow DARTS~\cite{liu2018darts} and transfer the cells searched on CIFAR-10 to ImageNet. Specifically, models are constructed by stacking 14 cells with 48 initial channels and are trained from scratch for 250 epochs by SGD with a 0.5 initial learning rate. Table~\ref{tab:compare_imagenet} shows that our transferred model achieves 76.6\% top-1 accuracy on validation set, outperforming all prior differentiable NAS methods. Additionally, due to the high efficiency of our method, we can also directly search on ImageNet. In the search stage, we follow DADA~\cite{dada} and randomly select 120 classes. A supernet is constructed by stacking 8 cells with 16 initial channels. We first warm up the operation weights and architecture parameters for 30 epochs and then jointly train architecture parameters and augmentation parameters for another 20 epochs.
As shown in Table~\ref{tab:compare_imagenet}, DAAS achieves 76.5$\%$ top-1 accuracy on ImageNet validation set.

\begin{table}[tb!]
\caption{Top-1 test error (\%) on CIFAR-10 of four architectures trained with default augmentation policies in prior NAS works and the policies discovered in our search space. Smaller value better performance. The performance ranking is given in bracket.}
\label{tab:compare_only_nas}
\centering
\resizebox{.95\columnwidth}{!}{
\smallskip\begin{tabular}{lccHcc}
\toprule
  &   \multirow{2}{*}{\textbf{Params} (M)}  &  \multicolumn{3}{c}{\textbf{Test Error} (\%)} & \multirow{2}{*}{$\Delta$} \\
  \cmidrule(lr){3-5}
  &  & NAS & Independent & \textbf{DAAS} & \\
\midrule
\textbf{Arch-1} & 4.40 & 2.58 (2) & & \textbf{2.09} (1) & $\downarrow$ 0.49  \\
\textbf{Arch-2} & 4.06 & 2.57 (1) & & \textbf{2.30} (3) & $\downarrow$ 0.27 \\
\textbf{Arch-3} & 4.12 & 2.64 (3) & & \textbf{2.36} (4) & $\downarrow$ 0.28 \\
\textbf{Arch-4} & 3.99 & 2.64 (3) & & \textbf{2.20} (2) & $\downarrow$ 0.44 \\
\midrule
\textbf{Average} & 4.14 & 2.61 & & 2.24 & $\downarrow$ 0.34 \\
\textbf{$\pm$std.} & $\pm$0.16 & $\pm$0.03 & & $\pm$0.10 & $\pm$0.07 \\
\bottomrule
\end{tabular}
}
\vspace{-10pt}
\end{table}

\begin{table*}[tb!]
	\caption{Ablation study on AA algorithm. We compare our PG method with prior works on CIFAR-10 and CIFAR-100 on various classic networks. The right block indicates the differentiable method for Auto-augmentation task, which are much faster than the reinforcement learning based method~\cite{AA}. Results are averaged over three parallel tests. $^\dagger$: results are obtained from~\cite{AA}. `-': results are not provided by the prior work. WRN and SS are the shorthand of Wide-ResNet and Shake-Shake, respectively.}
	\label{tab:comparison_aa}
	\centering
	\setlength{\tabcolsep}{1.0pt}
	\resizebox{\textwidth}{!}{
	\begin{tabular}{ll|cccc|ccccc}
	\toprule
	\textbf{Dataset} & \textbf{Model} & \textbf{Baseline}$^\dagger$ & \textbf{Cutout}$^\dagger$ & \textbf{AA}~\citeyearpar{AA} & \textbf{PBA}~\citeyearpar{pba} & \textbf{Fast-AA}~\citeyearpar{fast-aa} & \textbf{Faster-AA}~\citeyearpar{faster-aa} & \textbf{DADA}~\citeyearpar{dada} & \textbf{DDAS}~\citeyearpar{ddas} & \textbf{PG} (ours) \\
	\midrule
	CIFAR-10 & WRN-40-2 & 5.3 & 4.1 & 3.7 & - & 3.6 & 3.7 & 3.6 & - & \textbf{3.5}  \\
	CIFAR-10 & WRN-28-10 & 3.9 & 3.1 & \textbf{2.6} & 2.6 & 2.7 & \textbf{2.6} & 2.7 & 2.7 & \textbf{2.6} \\
	CIFAR-10 & SS(26 2x32d) & 3.6 & 3.0 & \textbf{2.5} & 2.5 & 2.7 & 2.7 & 2.7 & - & 2.7 \\
	CIFAR-10 & SS(26 2x96d) & 2.9 & 2.6 & 2.0 & 2.0 & 2.0 & 2.0 & 2.0 & 2.0 & \textbf{1.8} \\
	CIFAR-10 & SS(26 2x112d) & 2.8 & 2.6 & \textbf{1.9} & 2.0 & 2.0 & 2.0 & 2.0 & - & \textbf{1.9} \\
	CIFAR-10 & PyramidNet & 2.7 & 2.3 & \textbf{1.5} & \textbf{1.5} & 1.8 & - & 1.7 & - & 1.6 \\
	\midrule
	CIFAR-100 & WRN-40-2 & 26.0 & 25.2 & \textbf{20.7} & - & \textbf{20.7} & 21.4 & 20.9 & - & 21.0 \\
	CIFAR-100 & WRN-28-10 & 18.8 & 18.4 & 17.1 & 16.7 & 17.3 & 17.3 & 17.5 & \textbf{16.6} & 16.9 \\
	CIFAR-100 & SS(26 2x96d) & 17.1 & 16.0 & 14.3 & 15.3 & 14.9 & 15.0 & 15.3 & 15.0 & \textbf{14.6} \\
	CIFAR-100 & PyramidNet & 14.0 & 12.2 & \textbf{10.7} & 10.9 & 11.9 & - & 11.2 & - & 11.0 \\
	\bottomrule
	\end{tabular}
	}
	\vspace{-10pt}
\end{table*}

\textbf{Superiority of the searched augmentation policies.}
Prior differentiable NAS methods~\cite{liu2018darts,dong2019searching,rome} manually design and fix default data augmentation policies for all networks. To verify the superiority of our joint searching framework against the NAS framework, we first search on our joint search space and then train the discovered architectures with default data augmentation policies and our searched policies for 600 epochs. The results of four parallel tests are reported in Table~\ref{tab:compare_only_nas}, showing that our discovered augmentation policies can significantly improve the performance for all architectures. Specifically, the top-1 test error is reduced up to nearly 0.5\% and 0.34\% on average. 

We observe a significant difference in the performance ranking for the four architectures under default augmentation policies and our discovered policies, implying that it should consider data augmentation when evaluating the discovered architectures in NAS. Additionally, though the architectures perform similarly under the default policies, they show disparity once trained by proper augmentation policies, which helps distinguish better candidate architecture.

\subsection{Ablation Study}
\label{subsec:ablation_study}
\textbf{Effectiveness of AA algorithm.}
Table~\ref{tab:comparison_aa} compares our policy gradient based method (PG) with prior AA methods on CIFAR-10 and CIFAR-100 on multiple classic CNN networks: Wide-ResNet~\cite{wideResNet}, Shake-Shake~\cite{shakeshake}, and PyramidNet~\cite{pyramid}. After the search process, we follow AA~\cite{AA} and DADA~\cite{dada} by training Wide-ResNets for 200 epochs, Shake-Shakes for 1,800 epochs, and PyramidNets for 1,800 epochs. Our results are averaged over three parallel tests. Table~\ref{tab:comparison_aa} shows that our AA algorithm (PG) outperforms peer differentiable AA methods and even outperforms AA~\cite{AA} on 4 benchmarks, whose search cost requires thousands of GPU-days while ours is fewer than 1 GPU-days. 

\begin{table}[tb!]
\caption{Top-1 test error (\%) of joint and independent search for NAS and AA on CIFAR-10. `PG' indicates our AA method based on policy gradient algorithm. In the independent search mode, we first search four architectures and then search augmentation policies for them by DADA and our PG method.}
\label{tab:comparison_independent_search}
\centering
\resizebox{.98\columnwidth}{!}{
\smallskip\begin{tabular}{lccc}
\toprule
 &  \multicolumn{2}{c}{\textbf{Independent Search}} & \textbf{Joint Search} \\
 \cmidrule(lr){2-3} \cmidrule(lr){4-4}
  & NAS+DADA & NAS+PG & DAAS (ours) \\
\midrule
\textbf{Test-1} & 2.21 & 2.25 & 2.09 \\
\textbf{Test-2} & 2.39 & 2.32 & 2.30 \\
\textbf{Test-3} & 2.49 & 2.31 & 2.36 \\
\textbf{Test-4} & 2.65 & 2.63 & 2.20 \\
\midrule
\textbf{Avg. $\pm$ std.} & 2.44$\pm$0.18 & 2.38$\pm$0.17  & \textbf{2.24$\pm$0.10} \\
\bottomrule
\end{tabular}
}
\vspace{-10pt}
\end{table}

\textbf{Comparison between joint searching and independent searching for NAS and AA.}
To further verify our analysis on the strength of joint searching in Sec.~\ref{subsec:discussion}, we compare with independent searching for NAS and AA. For the settings of independent search, we first search architectures by alternately training operation weights and architecture parameters for 50 epochs and derive the final network according to the discovered cells. Then, we search augmentation policies by training augmentation parameters for another 50 epochs with a fixed architecture. Four parallel tests are conducted for both joint searching and independent searching, and the average performance is reported in Table~\ref{tab:comparison_independent_search}, showing that joint searching outperforms independent searching by 0.14\% (NAS+PG) and 0.2\% (NAS+DADA). 
However, unlike the independent searching scheme with two separate stages where the network architecture is fixed when searching augmentation policies, joint searching is more efficient and can adjust architectures and policies simultaneously.

\begin{table}[tb]
\caption{Top-1 test error (\%) of mixed combinations of architectures and policies on CIFAR-10. We randomly mix up the architecture and policies discovered in three parallel tests.}
\label{tab:ablation_study_arch_policy}
\centering
\resizebox{.75\columnwidth}{!}{
\smallskip\begin{tabular}{l|ccc}
\toprule
  &  \textbf{Policy-1}  &  \textbf{Policy-2} & \textbf{Policy-3} \\
\midrule
\textbf{Arch-1} & \textbf{2.09} & 2.29 & 2.37 \\
\textbf{Arch-2} & 2.56 & \textbf{2.30} & 2.31 \\
\textbf{Arch-3} & 2.37 & 2.31 & \textbf{2.20} \\
\bottomrule
\end{tabular}
}
\vspace{-10pt}
\end{table}
\textbf{Ablation study by mixing up the discovered architectures and policies.}
We show the validity of the discovered combinations of architecture and policies. We randomly mix up three discovered combinations and full train the mixed combinations for 600 epochs on CIFAR-10. The results are reported in Table~\ref{tab:ablation_study_arch_policy}, showing that architectures achieve the best performance when the related policies are utilized.

\section{Conclusion}
This work proposes an efficient differentiable joint search algorithm named DAAS to simultaneously search for efficient architecture and augmentation policies and constructs a bi-level optimization for joint search. Specifically, we introduce a policy gradient based second-order approximation to train augmentation parameters and adopt the Gumbel technique to train architecture parameters.
Extensive experiments and ablation studies verify the effectiveness of our method. In particular, DAAS achieves 97.91\% accuracy on CIFAR-10 and 76.6\% top-1 accuracy on ImageNet in only 1 GPU-days' search.
Also, this work shows the superiority of joint search for AA and NAS, implying that NAS evaluation should consider appropriate data augmentation policies.

\bibliography{mybib}
\bibliographystyle{icml2022}



\end{document}